\documentclass[11pt, a4paper, logo, twocolumn, internal, copyright, nonumbering]{googledeepmind}

\usepackage[authoryear, sort&compress, round]{natbib}
\usepackage{subcaption}
\usepackage{graphicx}

\title{Wanting to be Understood}

\correspondingauthor{chrisantha@google.com}

\paperurl{arxiv.org/abs/123}

\reportnumber{001} 

\renewcommand{\today}{April 2025}

\author[1]{Chrisantha Fernando}
\author[1]{Dylan Banarse}
\author[1]{Simon Osindero}

\affil[1]{Google DeepMind}

\begin{abstract}
This paper explores an intrinsic motivation for mutual awareness, hypothesizing that humans possess a fundamental drive to understand \textit{and to be understood} even in the absence of extrinsic rewards. Through simulations of the perceptual crossing paradigm, we explore the effect of various internal reward functions in reinforcement learning agents. The drive to understand is implemented as an active inference type artificial curiosity reward, whereas the drive to be understood is implemented through intrinsic rewards for imitation, influence/impressionability, and sub-reaction time anticipation of the other. Results indicate that while artificial curiosity alone does not lead to a preference for social interaction, rewards emphasizing reciprocal understanding successfully drive agents to prioritize interaction. We demonstrate that this intrinsic motivation can facilitate cooperation in tasks where only one agent receives extrinsic reward for the behaviour of the other.
\end{abstract}

\begin{document}

\maketitle

\section{Introduction} 

\noindent Evidence from developmental psychology and primatology suggests that humans (and possibly other primates) have a drive to establish new forms of \emph{shared intentionality} \citep{tomasello2022evolution}. Here, ``shared intentionality'' refers to the capacity for two or more individuals to align their intentions, goals, values, policies, and world models in a mutually recognizable way.

\noindent This drive aims to achieve \emph{common knowledge} of each other, eventually resulting in understanding the other through the intentional stance \cite{dennett1989intentional} and in more bodily ways \citep{dipaolo2008perceptual}\citep{clark1998extended}\citep{varela1991embodied}. In other words, we struggle afresh in each lifetime to develop techniques of \emph{primary intersubjectivity} \citep{trevarthen1979communication}, which are basic methods for relating to one another. Primary intersubjectivity typically appears in infancy and includes face-to-face interactions, mutual gazes, and the early emotional attunement between caregiver and child.

\noindent Later, we develop \emph{secondary intersubjectivity} \citep{vygotsky2012thought}, expanding our interactions to include a third entity in a shared context. This step involves joint attention, referential gestures or performances, and eventually language. Crucially, these activities are often pursued \emph{purely for the sake of establishing shared understanding}, rather than for any immediate extrinsic task rewards \citep{moeller2023human} or for conforming to externally imposed ``language games'' \citep{lewis1969convention}.

We hypothesise that through the Baldwin
effect\footnote{The Baldwin effect describes a mechanism by which traits initially acquired through individual learning within a lifetime can eventually become genetically encoded over evolutionary time through selection for mutations that facilitate faster learning, and eventually leading to their innate emergence.}
\citep{baldwin1896new} operating during the Paleolithic in the context of diverse extrinsically rewarded social coordination tasks \citep{fernando2024origin} a genetic intrinsic motivation evolved that rewarded the mutual creation of social coordination ``for the sake of it'' in a new way which had not been an intrinsic motivation for apes \citep{tomasello2022evolution}. Such minimal 
pre-Gricean\footnote{Gricean communication involves inferences about speakers' intentions (hidden mental states) and shared assumptions (conversational maxims). Pre-Gricean systems represent an earlier evolutionary stage, involving more basic signaling \citep{moore2017gricean}}\citep{scott2015speaking} ``cognitive gadgets'' \citep{heyes2018cognitive} allowed the bootstrapping of language, art, and open-ended culture. We propose humans evolved a motivation not merely to understand, but also to be understood by the other. That is, to want the other to understand one's values, goals, policies, and world models, in short, it brought about an innate struggle for mutual awareness, promoting the establishment of primary and secondary intersubjectivity \citep{trevarthen1979communication} which allowed shared intentionality \citep{tomasello1993cultural} \citep{tomasello2005understanding} manifested in turn-taking \citep{sacks1974turn}, imitation \citep{meltzoff1989imitation}, cultural evolution \citep{boyd1985culture}, language \citep{tomasello2008origins}, and other varieties of external representation making such as art and music \citep{noe2023entanglement}.

In fact, such a core set of cognitive gadgets may explain why "the hard problem of 
consciousness" is a problem at all \citep{chalmers1997conscious}. 
The hard problem of consciousness refers to the challenge of 
explaining to another why and how physical processes in my brain give rise to subjective, qualitative experiences (qualia) – the feeling of what it's like for me to be a conscious organism.
We are troubled by the fact that due to bandwidth restrictions \citep{mcgregor2005levels}, our external representations of our raw experience of red, \emph{qualia},
will never be able to convey the way it truly feels, or why it feels that way to me. There is an \textit{a priori} impossibility of satisfactory explanation either to myself or to another why red feels as it does to me \citep{Jackson1982}, but why do we keep on trying to make the other understand? \\

Exactly how, if such a drive for wanting one's raw experience to be understood, can being-understood-by be measured by the sender in the absence of extrinsic reward of the type provided in language games specified top-down \citep{lewis1969convention}? In Lewis language games of the type introduced by Wittgenstein \citep{wittgenstein1953philosophical}, communicative success is externally defined, and reward is given to both agents symmetrically \citep{steels1997synthetic} \citep{steels1999talking}. We would like such an intrinsic motivation to explain the open-ended unsupervised invention of new language games, for the sheer joy of it. \\

Any such intrinsic motivation must be biologically plausible; it cannot depend on direct access to the actions and brain states of another agent but must use internal models of the other agent, these being a function only of one's observations, actions, predictions, and one's own brain states \citep{jaques2019socialinfluenceintrinsicmotivation}. But unlike earlier work, we propose that 1. such a drive should not require sophisticated counterfactual reasoning, as it appears to exist shortly after birth \citep{hobson2004cradle}, and we explore the possibility 2. that the drive is not only a drive to influence, but a drive to be influenced by another. In \cite{jaques2019socialinfluenceintrinsicmotivation} whilst communication protocols emerged, they did so because of an extrinsic reward shared by all agents, i.e. the need for shared intensionality was pre-supposed by the task, the task was not established for the sake of shared intensionality per se. \\

A necessary (but not sufficient) condition for the drive we seek is that when incorporated into an agent it results in the agent establishing a high interaction equilibrium with other agents with the reciprocal drive in the absence of extrinsic reward \citep{lerique2022embodied}. This corresponds to the concept of primary intersubjectivity defined by \cite{trevarthen1979communication}. Another much more stringent requirement is that this intrinsic motivation results in agents co-inventing a set of external representations such as performances, signs or symbols that are \textit{about} other entities in the world \citep{fernando2024origin}, i.e. secondary intersubjectivity utilizing techniques for joint attention such as pointing and other forms of reference \citep{trevarthen1978secondary} which allows shared intentionality, e.g. joint goals. \\

\begin{figure*}[h!]
\includegraphics[width=17cm]{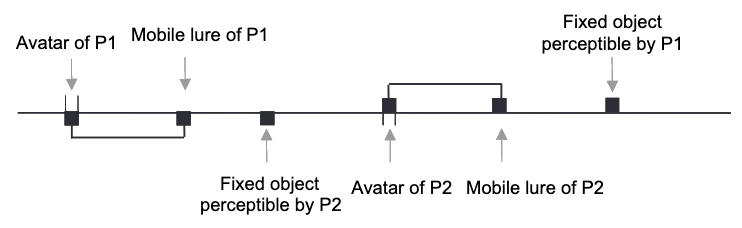}
\caption{\footnotesize The perceptual crossing paradigm, Figure from \cite{auvray2009perceptual}. Two agents (avatars) live on a line. They can cross each other, each other's shadows, or their own private stationary object.} 
\label{PCP}
\end{figure*}

We focus in this paper purely on potential drives for primary intersubjectivity, and study them within a simulation of a minimal experimental paradigm which has been used extensively in humans, namely, the perceptual crossing paradigm (PCP) \citep{auvray2009perceptual}, see Figure \ref{PCP}. In the PCP, two humans interact using only a mouse which controls an avatar which can move around on a circle. When their avatars cross,  both perceive a buzz. If an avatar crosses their private stationary object on the circle they also feel a buzz. In addition, each avatar has a `shadow' which is situated a fixed displacement from themselves. If an avatar crosses the other's shadow, they feel a buzz but the other does not feel a buzz. \\

In the original experiment, subjects were asked to press a button when they thought they were interacting with the other. Human pairs spend most of their time interacting with each other in preference to the fixed object or each other's shadows. They develop turn-taking strategies to ident ify each other reliably, and some pairs of humans play games of dance-like anticipation with each other \citep{declerck2009rendre} involving passive (following) vs active (leading) turn-taking \citep{kojima2017sensorimotor}.\\

Previous simulations of the PCP evolved a policy within an evolutionary robotics framework \citep{iizuka2007minimal}, \citep{dipaolo2008perceptual}, the fitness function being to be close to the other agent or to click when close \citep{froese2010modelling}, \citep{auvray2012perceptual}. Whilst this resulted in the number of self-other crossings increasing over evolutionary generations, the resulting policy does not constitute a valid intrinsic motivation because it presupposes an oracle which can determine that one is interacting with another agent. In this paper it is the nature of such an `other' detecting `oracle' which interests us. The policies that evolutionary robotics approaches produce in such simulations are instead exquisitely over-fitted to this particular task, and will not generalize to other tasks. Instead, we wish to propose intrinsic motivation functions which do have the potential to show preference for self-other interaction without \textit{a priori} being told that an interaction with another agent is occurring. \\


\section{Empirical Evidence} 

Imagine placing a handful of newborn infants on a desert island with no access to humans, their life support provided by soulless robots. Would they develop language? If so then some genetic information (plus information from the desert island) must account for why humans do so (and chimpanzees do not). If they do not develop language within one generation, then some information transmitted, not genetically, but culturally, is needed to bootstrap language, such as a tradition of prior language use. For example, a transformer trained with a supervised next token prediction loss such as ChatGPT \citep{openai2023gpt4} will not reinvent language if not exposed to language data, but it will admirably pass on the tradition of at least the most effortless aspects of language use. In this sense ChatGPT may be said to understand language, but to not want to be understood enough to go to the trouble of inventing new language \citep{rita2022emergent} \citep{lazaridou2020emergent}. \\

Tomasello has proposed that humans differ from apes in that humans are intrinsically motivated to achieve shared intentionality, i.e. to form joint goals, joint intentions, joint attention, create mutual knowledge and form shared beliefs (p7 \citep{tomasello2008origins}), using prosocial motivations and norms for helping and sharing. Collaboration is intrinsically rewarding (p178, \textit{ibid}) sometimes more rewarding than the instrumental goal of the task itself \citep{warneken2006cooperative}. In short, children collaborate for the sake of collaborating. But as described by Kaplan and Hafner ``Tomasello remains relatively elusive on what exactly this ``sharing motivation'' consists of. How does the brain recognize ``shared experiences''? What is special in situations of ``joint intentionality''? (Sect. 3.4 \citep{kaplan2006challenges}). \\

A drive for primary intersubjectivity is supported by Trevarthen's observations of infant coordinative gesture use, vocalizations, and affect, in face-to-face protoconversations with caregivers \citep{trevarthen1979communication}. Primary intersubjectivity involves dyadic interactions without reference to a third object or person. Travarthen was concerned with how 2-3 month infants and mothers came to share control of joint sensorimotor contingencies (SMCs) and how they could predict what the other would do and know, in quite a different way to how SMCs are established with inanimate objects. For example, intersubjective behavioural dynamics include synchronization and alternation between infant and mother. The mother makes attempts to attract the infant's attention, principally to obtain eye contact, if the baby makes an expression the mother mirrors it, and when the baby is ready to make an utterance the mother `steps down'. Watching, admitting, withdrawing, makes the mother subordinate to the infant's communicative intentions. A complex form of mutual understanding develops. \\

Stern describes how during the first months of life infants orient more to highly predictable SMCs from self-generated events, but after 3 months to more imperfect SMCs (other-generated events) \citep{stern2018interpersonal}. More recent work shows that children establish dyadic SMCs even earlier, aligning their gaze with the timing of adults’ vocal behaviors \citep{crown2002cross}, and by 2 months raising their arms in anticipation of being picked up \citep{reddy2013anticipatory}. Infants of only 45 minutes of age make strenuous efforts to copy an adult's tongue protrusion and match other facial expressions and vocal rhythms, and at later ages seem to try to infer the goals of adult demonstrators \citep{hobson2004cradle} \cite{kugiumutzakis2015neonatal} \citep{meltzoff1989imitation}. The ``still-face-paradigm'' shows infants become extremely distressed when their caregiver suddenly holds a neutral face \citep{tronick1978infant} or the caregivers responses are delayed in video feedback by only 30 seconds \citep{murray1985emotional} \footnote{We do not consider secondary intersubjectivity here. A drive for secondary intersubjectivity may be needed to explain the special kinds of triadic interaction between two humans and an object, e.g. joint attention involving informational helpful pointing, giving of spontaneous help such as picking up dropped objects, involving alignment with others' goals and attitudes to objects without direct external rewards \citep{warneken2006altruistic}, organizing collaborative activity with role taking and reversal \citep{warneken2007helping}, and perspective taking during cooperation\cite{}. Chimp gestures are limited to imperatives (asking for things) and getting attention, and although they can be learned in pairs, they are not transmitted culturally. They do not 'get it' that anyone would want to help them by giving them cooperative information declaring knowledge. They do not ask questions to get answers (p72, \citep{tomasello2008origins}). In the perceptual crossing experiment there is absolutely no scope for secondary intersubjectivity (by design) because there is no third object that both agents can interact with, and therefore, secondary intersubjectivity cannot be studied with this task}.\\

Whilst the desert island baby experiment is a Gedankenexperiment, some poor approximations to it do exist, and suggest that there is an intrinsic drive for establishing communicative relationship. Firstly, profoundly deaf children raised without sign language, out of desperation, will invent a language together called homesign \citep{sacks2009seeing}. Confounders here are that these children can observe full language users, and their external representations, and be scaffolded also by the pre-linguistic interactional drives of existing language users. \\

Finally, in autism, mutual engagement with neurotypicals my fail \citep{milton2012double}. From the neurotypical perspective, there can sometimes be a sense that one is being treated as an inanimate object rather than as a person \cite{shanahan2024simulacra}, and from the autistic perspective one may be able to coordinate better with other autistic individuals than neurotypicals \citep{heasman2019neurodivergent}. One reason this could be is that there is misalignment in methods that are being used to establish mutual awareness, and therefore expectations of what an engagement should constitute \citep{bolis2018beyond}. But precisely what is responsible for this mismatch of engagement styles? \\

\section{Methods} 

We examine a multi-agent interaction scenario in a 1D ``Perceptual Crossing'' environment in which two agents each occupy a position on a circular or modular 1D line (0 to 1, wrapped) \cite{auvray2009perceptual}. Each agent can move left, move right, or remain still, attempting to elicit and anticipate “crossing” events. A crossing occurs whenever an agent’s position coincides with a salient stimulus within a predefined distance threshold—such as another agent, the other agent’s “shadow” (the position offset by 0.2 modulo 1), or a fixed (stationary) object. Each agent receives a step-based observation that includes its current position, the previous step’s crossing value, its last action, and its last received reward. Both agents select discrete actions from the set {left,right,no-op}, which affect their positions by a fixed increment of 0.05 or no movement. We train these agents jointly using a reinforcement learning (RL) framework \citep{Liang2018}. \\

Separately from the main RL policy, each agent also runs a small Long-Short Term Memory (LSTM) recurrent neural network \citep{Hochreiter1997} that predicts whether the next time step will yield a crossing or not. This recurrent predictor is trained offline from trajectories collected over episodes: after each episode, the agent’s sequence of observations and actual crossing values is appended to a dataset, which is then used to update the LSTM model with supervised learning. At each environment step, the trained LSTM outputs a predicted crossing and an associated confidence, which can be used to provide additional shaping rewards or to detect “uncertainty” states.\\

We use Proximal Policy Optimization (PPO) \citep{Schulman2017} via the Ray RLlib library to train both agents simultaneously \citep{Moritz2018}. Each agent’s policy network is an LSTM-based model receiving the 4D observation described above; training proceeds in episodic form, with one episode lasting a fixed number of 100 steps. We now describe the various reward functions we investigate. 
\subsection{A Drive For Understanding}

\begin{figure}[h!]
\includegraphics[width=7cm]{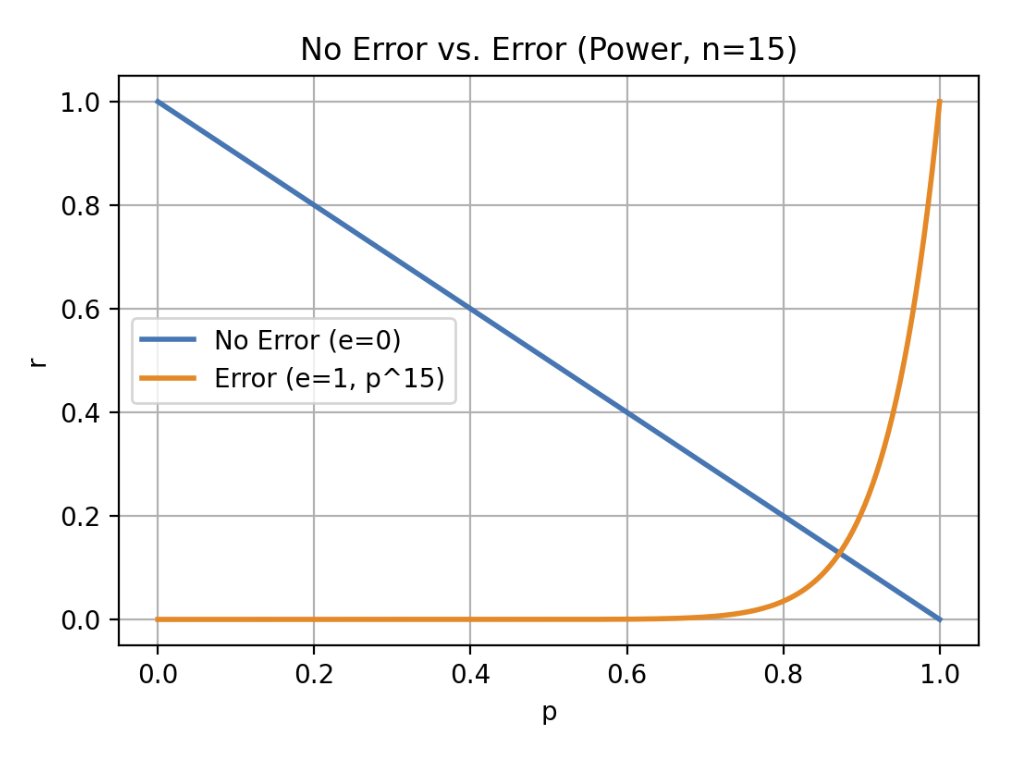}
\caption{\footnotesize The artificial curiosity reward rewards policies that minimize errors when certainty is low, but maximizes errors when certainty is high. This is analogous to conducting scientific experiments by setting up a situation in which a prediction is strong, and attempting to contradict that prediction. Reward is shown on the y-axis, and certainty on the x-axis.  } 
\label{artifical_curiosity_reward}
\end{figure}

Intrinsic rewards are mechanisms to encourage specific behaviors from each agent. As described above, each agent maintains an LSTM which predicts whether a crossing will occur in the next timestep and outputs a confidence \(p(c)\).The drive for understanding is modelled in the form of the reward function shown in Figure \ref{artifical_curiosity_reward}. 


\begin{equation}
r(e, p) = \begin{cases} 
    1 - p & \text{if } e = 0 \\
    p^n & \text{if } e = 1 \\
\end{cases}
\end{equation}

If an agent's predictor makes no error then the policy is rewarded in proportion to the uncertainty of that prediction. If the agent's predictor makes an error however, then there is an exponentially increasing reward for this error as the certainty of the prediction reaches a probability of 1. This means the policy is rewarded, once the predictor becomes quite good, for spending time in regions of the environment which are expected to be predictable but can continue to be surprising. Is such surprise sufficient to make agents prefer to interact with each other as proposed elsewhere \citep{lenay2011you} \citep{friston2024supervised}? 

\subsection{A Drive For Being Understood}

A drive for being understood can be implemented in various ways. We consider three interpretations of what it means to be understood. 

\subsubsection{To Imitate and Be Imitated-By} 

This intrinsic motivation mechanism encourages imitation and being imitated by another. This has been previously studied in the context of simultaneous extrinsic rewards \citep{eccles2019learningreciprocitycomplexsequential} in which the intrinsic motivation was to minimize the difference between the objectively determined behavioural trajectories of both agents. Using such a measure of what it means to imitate, but without a concomitant extrinsic reward as they used, has the unfortunate null behavioural attractor where both agents can maximize reward by doing nothing. To avoid this, we implement a reward signal based on the observation of reciprocal action patterns within the agents' recent interaction history. Unlike \cite{eccles2019learningreciprocitycomplexsequential} two distinct and reciprocal imitation reward functions are utilized: promote imitation and imitate.\\

Promoting Imitation: This function rewards an agent when it actively creates a certain pattern of crossings by its own active actions (actions 1 or 2) in the initial segment of its recent history, which is then followed by a passively observed set of crossings which are the same as it produced, consistent with the possibility that the other agent has imitated its recent behavior. This encourages an agent to initiate actions that will then be followed by the other agent's imitation of them. \\

Imitation Reward: Conversely, the imitation function rewards an agent for passively observing a pattern in the initial segment of its recent history (i.e. whilst doing action 0) followed by actively replicating that pattern using actions 1 or 2 in the subsequent segment. This function rewards an agent for following the other agent's lead.\\

Both functions analyze the recent interaction history of each agent, tracking actions and crossing events. A reward is generated proportionally to the frequency of observed reciprocal action patterns, normalized by the number of opportunities to observe those patterns. The history array is 10 steps long, and the first 5 steps are compared to the second 5 steps. Only crossing transitions (0->1, 1->0) are compared in these two windows, not absolute crossing states.\\

\subsubsection{Influence and Impressionability} 


\begin{figure}[h]
    \centering
    \begin{subfigure}{0.40\textwidth}
        \centering
        \includegraphics[width=7cm]{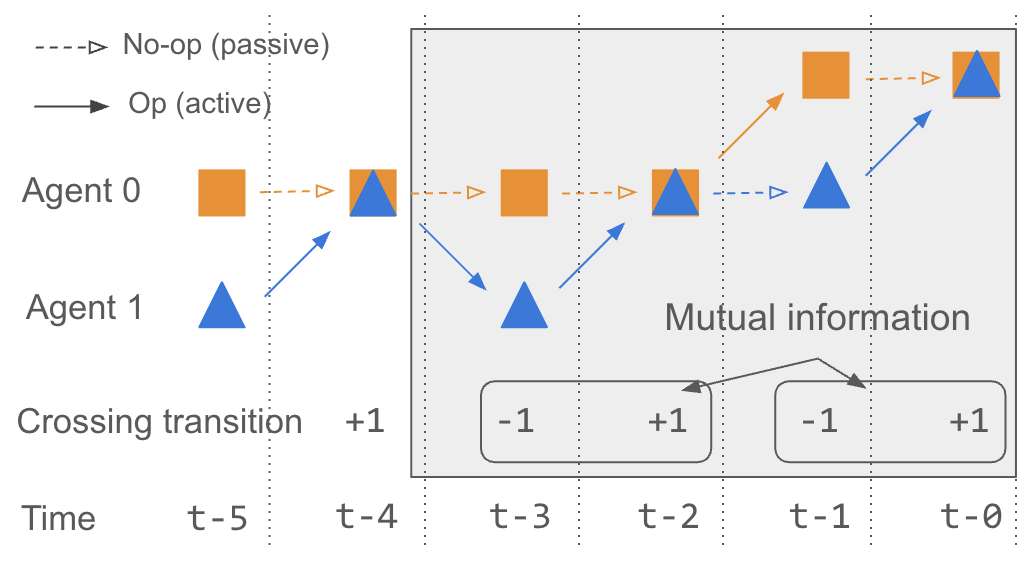}
        \caption{\footnotesize Agent0 is passive for two steps followed by actively causing a crossing transition; recreating the crossing transitions observed in the passive phase.}
        \label{fig:sub_a}
    \end{subfigure}
    \hfill
    \begin{subfigure}{0.45\textwidth}
        \centering
        \includegraphics[width=7cm]{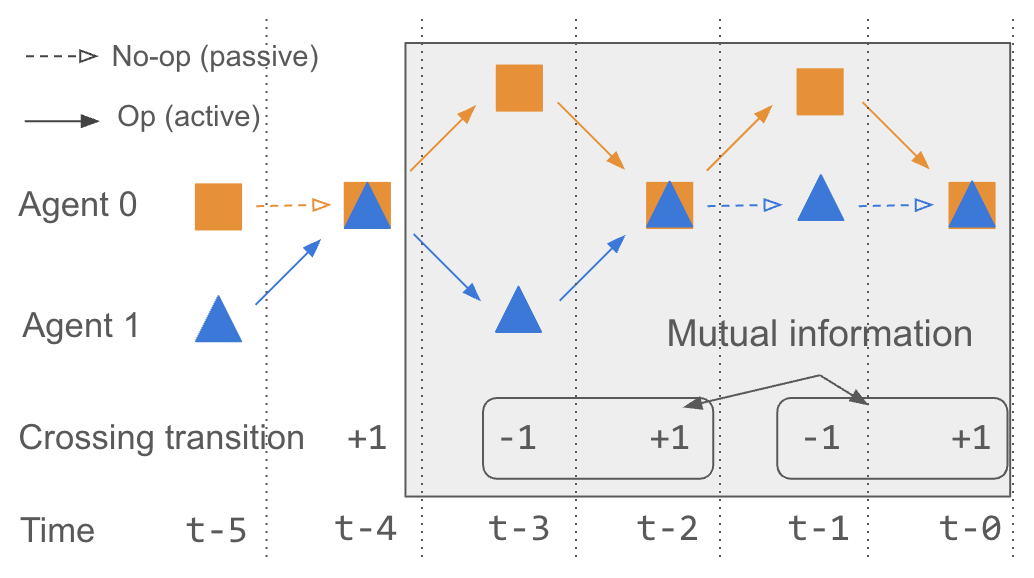}
        \caption{\footnotesize Agent1 is active and then passive, observing transitions caused by agent0.}
        \label{fig:sub_b}
    \end{subfigure}
    \caption{\footnotesize Mutual information is calculated (between two pairs of crossing transitions) when either of two interaction conditions are met. For passive-->active interactions an agent must do two no-ops, then at least one op with an associated crossing transition. For active-->passive interactions an agent must do at least one op associated with a crossing transition followed by two no-ops.}
    \label{mi_method}
\end{figure}

In a more general version of the above, we give each agent a reward for maximizing the recent contribution of the mutual information between their recent actions and the subsequent observations of the other, plus the mutual information between their observations of the other and their subsequent actions, see Figure \ref{mi_method}. Agents both want to be controlled and to control. In other words, agents want to both influence and be influenced by each other. This is similar to the imitation reward, except instead of requiring perfect imitation and imitation promotion, we relax the requirement and only require high mutual information between passively observed crossings and actively generated crossings, and vice versa. \\

To calculate their recent contribution to mutual information, each agent maintains a rolling buffer of the last 20 4-timestep chunks. At each timestep the observed crossings and action is stored. The mutual information between passively observed crossing events in the first 2 timesteps of the chunk and the actively produced crossing events in the last 2 timesteps of the chunk is calculated, over the whole rolling buffer of 20 chunks. Dirichlet priors are added (pseudo-counts in the frequency table of events used to calculate MI) in order to prevent exquisite sensitivity of reward to low counts, a factor which can considerably influence the shaping of the policy. This same calculation is then made over just the first 15 chunks in the rolling history. The Impressionability reward is calculated as the difference between the MI calculated over the full buffer and the MI calculated over a first 15 chunks of the buffer, to encourage the maximisation of change in mutual information. This is repeated in the reciprocal case of active crossings followed by passively observed crossings, and this is the Influence reward. Impressionability and Influence rewards are summed and this constitutes the full reward. This encourages the agents to influence each other, and be influenced by each other. Related work has rewarded agents for having causal influence over the environment \citep{seitzer2021causal}, but does not consider a symmetric influence of the environment on the agent. \\

Finally, we investigate the effect of removing one half of this drive, the drive to be impressionable, and leaving only the drive to influence. In this setting, both agents are trying to influence as in \cite{jaques2019socialinfluenceintrinsicmotivation} but have no intrinsic reward to be influenced by another. \\

\subsubsection{Sub-Reaction Time Anticipation} 

One sense in which I can know that you have understood me is that your expectations of what I will do next are correct, i.e that you can anticipate my actions before you could have had time to respond to them reactively. For example if there is a delay of 2 timesteps between your observation and your action, and you are able to maximize the mutual information between your observations of me and your responses to me, faster than in 2 timesteps, then this must mean your policy had, in an implicit sense, anticipated the observations two timesteps before your action, and produced your actions in anticipation to fit with those events before they could have been actually observed. \\

For example, if I say "Would you like to come ... " and you complete my sentence "for dinner", starting just after I say "co.." then I know that you have anticipated me asking you for dinner. Therefore I know what you know about me, by virtue of the fact that you have demonstrated this knowledge of what I was to say. If I am also rewarded for producing such anticipations in you, then I will act so as to allow you to anticipate me better. Therefore,  we simply add a 2 timestep action delay to both agents between their observation and their action, keeping the same ``Influence and Impressionability'' intrinsic motivation function above. \\

\subsubsection{Benefit of Intrinsic Drive to be Understood}

Finally we address the question, if the agents are required to complete a task in which cooperation is necessary, but only one agent gets an extrinsic reward, can the existence of intrinsic motivations for coordination help? Such a task is harder than a standard Lewis Language Game because there is no top down reward for both agents on completion. The task is as follows. Agent0 is given a new observation +1 or -1 randomly determined for each episode. If this signal bit is +1 then agent0 gets +1 units of reward for each timestep during that episode when agent1 is in the top half of the space (0.5 to 1.0) but gets punished by -1 for each timestep agent1 is in the bottom half (0.0 to 0.5) of the space. The opposite is true if the signal bit = -1. However, agent1 gets no reward for being anywhere, irrespective of the signal bit setting. Finally, we remove the Impressionability component of the mutual information reward and observe whether this interferes with the extrinsic reward task. 

\section{Results} 

\subsection{A Drive to Understand} 

\begin{figure*}
\includegraphics[width=16cm]{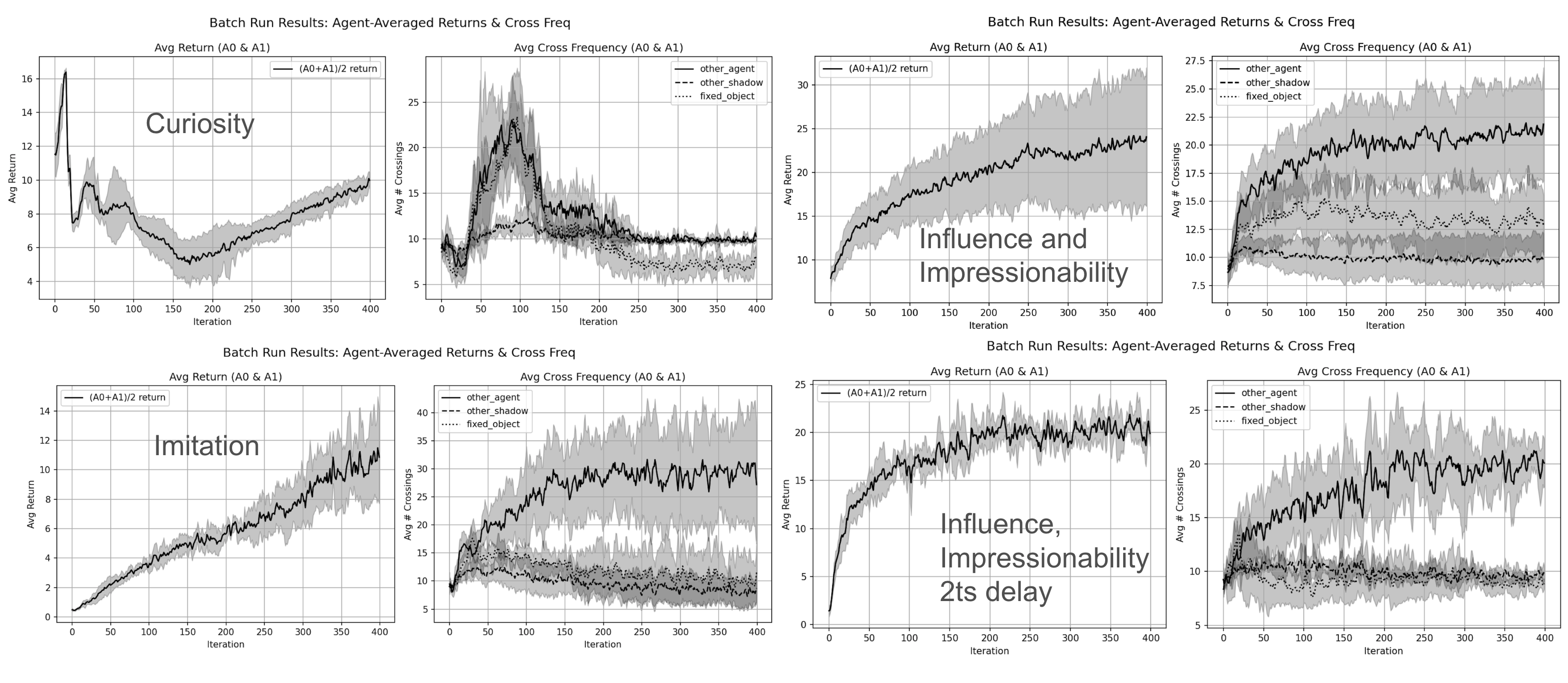}
\caption{\footnotesize Mean and standard deviation of rewards and crossing frequencies averaged over both agents over 5 runs.} 
\label{stats}
\end{figure*}

\begin{figure}
\includegraphics[width=7cm]{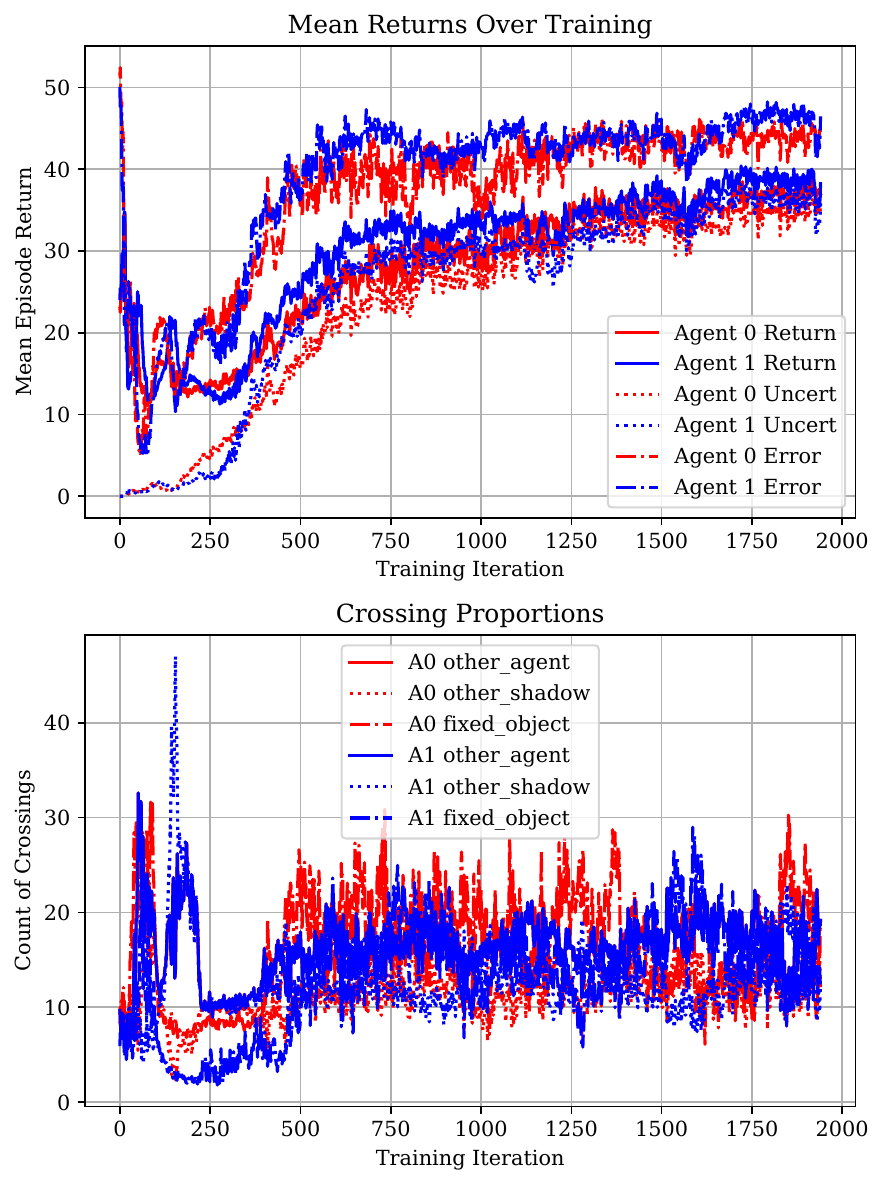}
\caption{\footnotesize Artificial Curiosity Reward: Artificial curiosity does not produce an overall preference for interaction with each other. Agents actively explore both  } 
\label{returns}
\end{figure}

In this control we consider only the drive to understand (artificial curiosity) and not the drive to be understood, see Figure \ref{artifical_curiosity_reward}. Figure \ref{stats} shows the mean and standard deviation of results averaged over 5 independent runs for each reward setting. The returns and crossing proportions over a typical developmental trajectory over 2000 episodes of experience is shown in Figure \ref{returns}. 

\begin{figure*}
\includegraphics[width=16cm]{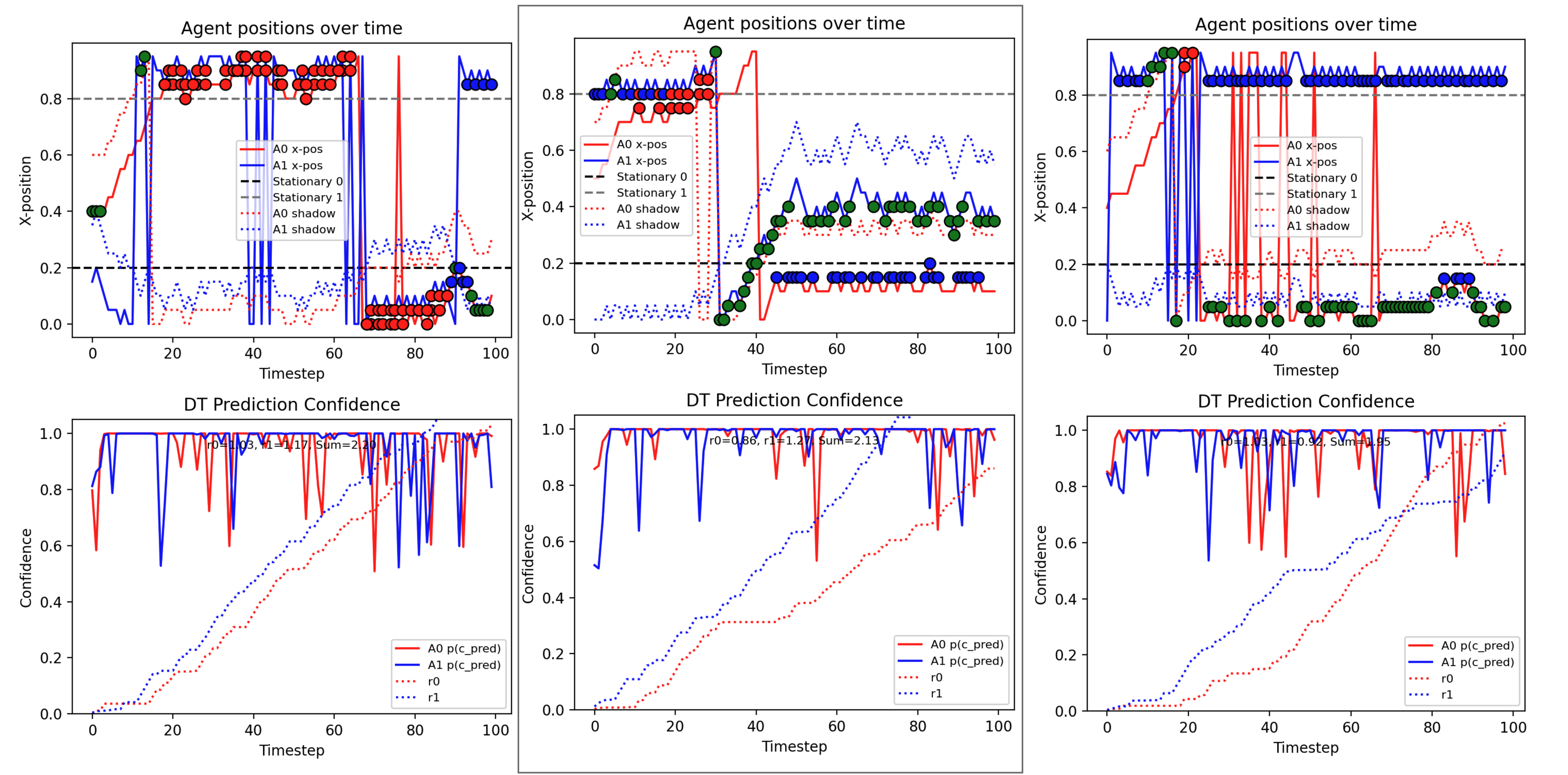}
\caption{\footnotesize Artificial Curiosity Reward: Three typical episodes from iteration 2100. Self-other crossing (red), self-shadow crossing (green), self-stationary crossing (blue). Cumulative reward (dotted lines on bottom graph) increases more when agents are in more unpredictable interactions e.g. with the shadow or with the other directly.}
\label{experimental_trajectories}
\end{figure*}

There is no preference for interaction with the other (solid blue line on the lower graph). Rewards are initially dominated by reducing errors, but quickly become dominated by creating errors in situations of certainty. There is a brief period around 200 iterations where self-other interactions are strongly preferred. Three typical episodes are shown in Figure \ref{experimental_trajectories} which shows that cumulative rewards increase more rapidly when agents are interacting with the other or the other's shadow. In all cases, agents like to make predictable crossings with an object, moving away from that object, and coming back to the object for repeated interactions with it. \\


\subsection{A Drive to Be Understood}

\subsubsection{To Imitate and Be Imitated By} 

\begin{figure}[h]
\includegraphics[width=7cm]{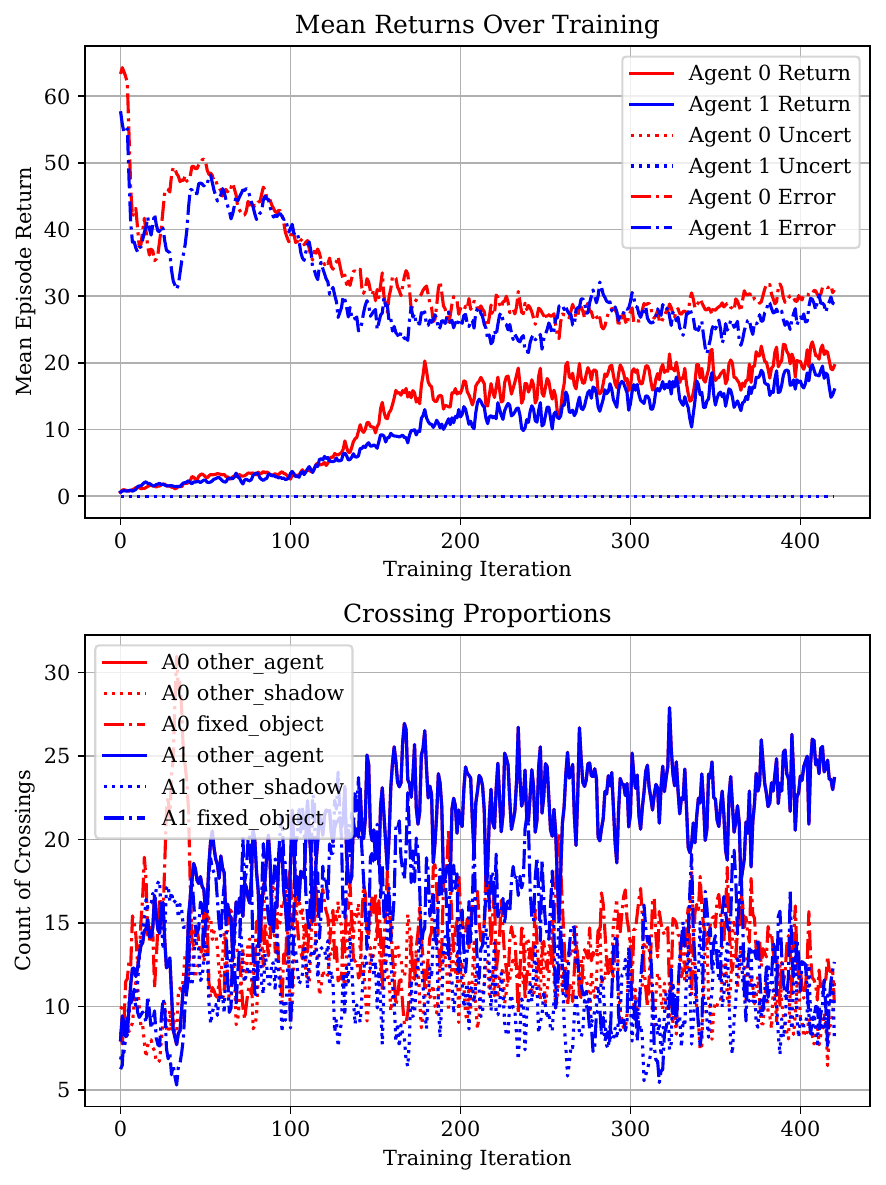}
\caption{Imitation: The thick blue line shows the frequency of self-other crossings far exceeds that of all other kinds of crossing with this intrinsic motivation.} 
\label{paper_plots_imitation}
\end{figure}

\begin{figure*}
\includegraphics[width=17cm]{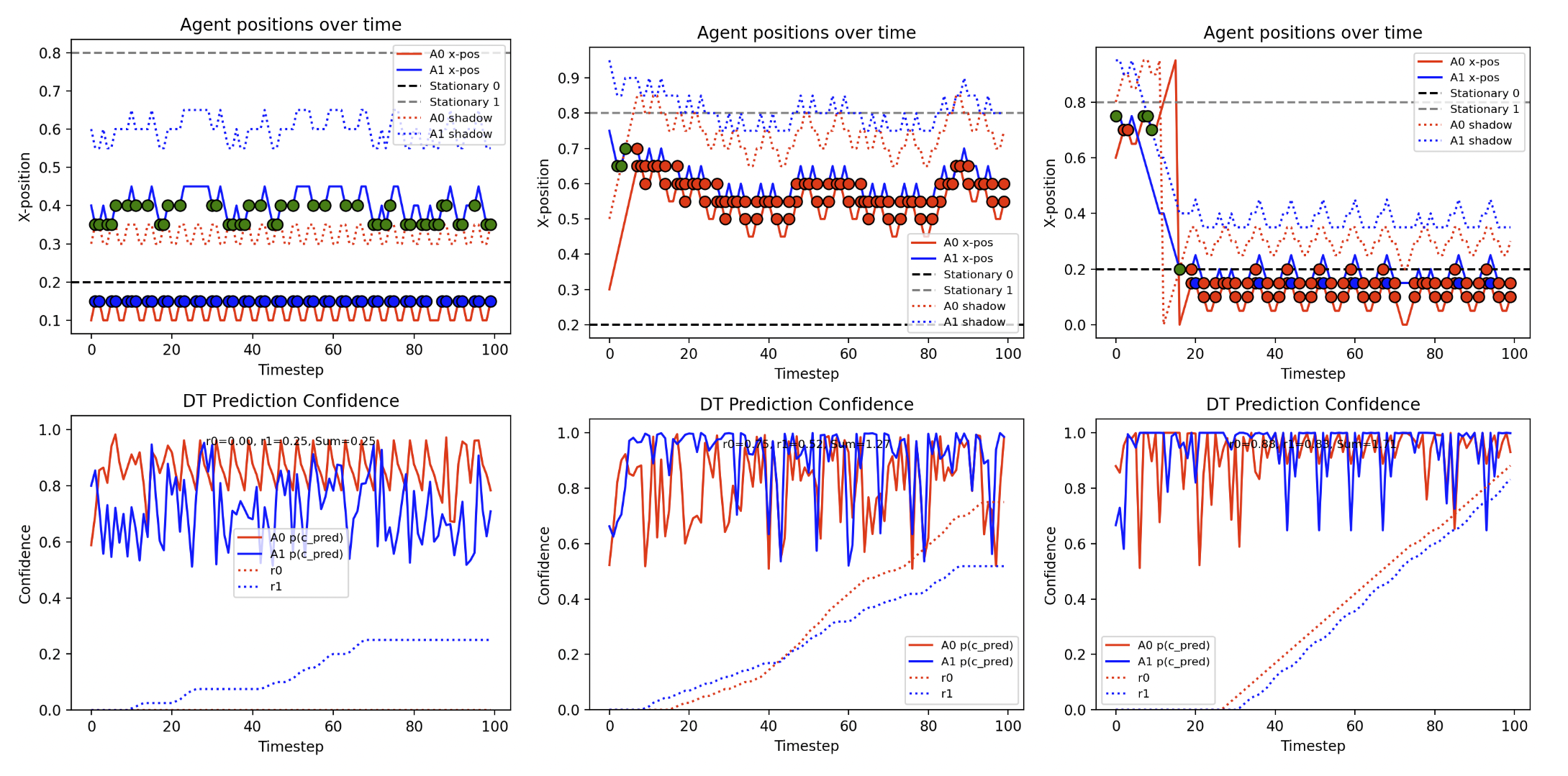}
\caption{\footnotesize Imitation: Over the course of the run, agents learn that most reward can be achieved by imitating and being imitated by with the other.} 
\label{imitation_trajectories}
\end{figure*}

We consider here only the drive to explicitly imitate and be imitated-by. Figure \ref{paper_plots_imitation} shows the returns and crossing proportions over a typical developmental trajectory, notably that self-other crossings predominate. Figure \ref{imitation_trajectories} shows a sample of 3 episodes from the end of the run. Self-other association consists of periods of repeated crossing and uncrossing with very stereotyped patterns. Limited reward can be achieved by crossing a shadow. Interacting with a stationary object gives no opportunity for reward. \\

\subsubsection{Influence and Impressionability} 

\begin{figure}
\includegraphics[width=7cm]{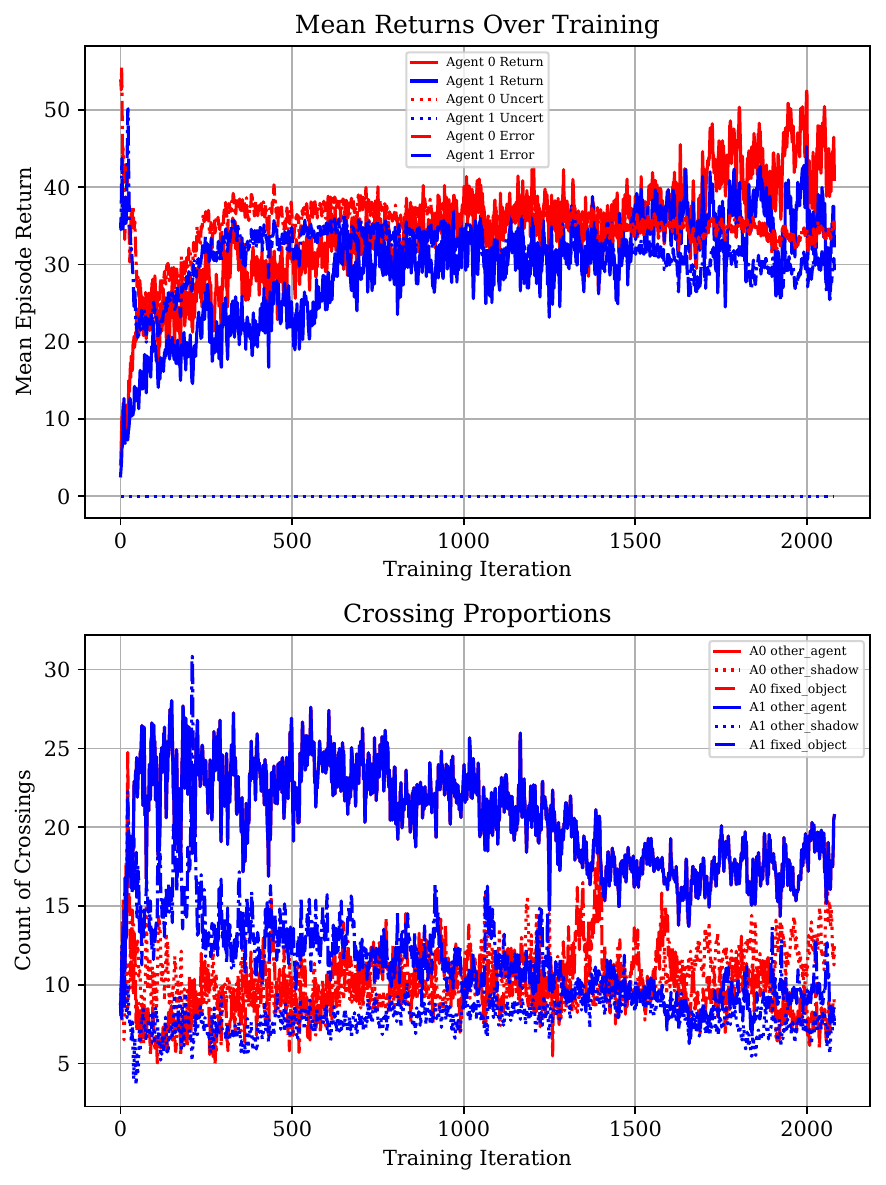}
\caption{\footnotesize Influence and Impressionability: The thick blue line shows the frequency of self-other crossings far exceeds that of all other kinds of crossing with this intrinsic motivation. However, the increase in reward is not monotonically related to the number of self-other crossings which actually decreases by 1500 iterations as the episode returns are increasing for both agents.} 
\label{paper_plots_mi_reward3}
\end{figure}

\begin{figure}
\includegraphics[width=7cm]{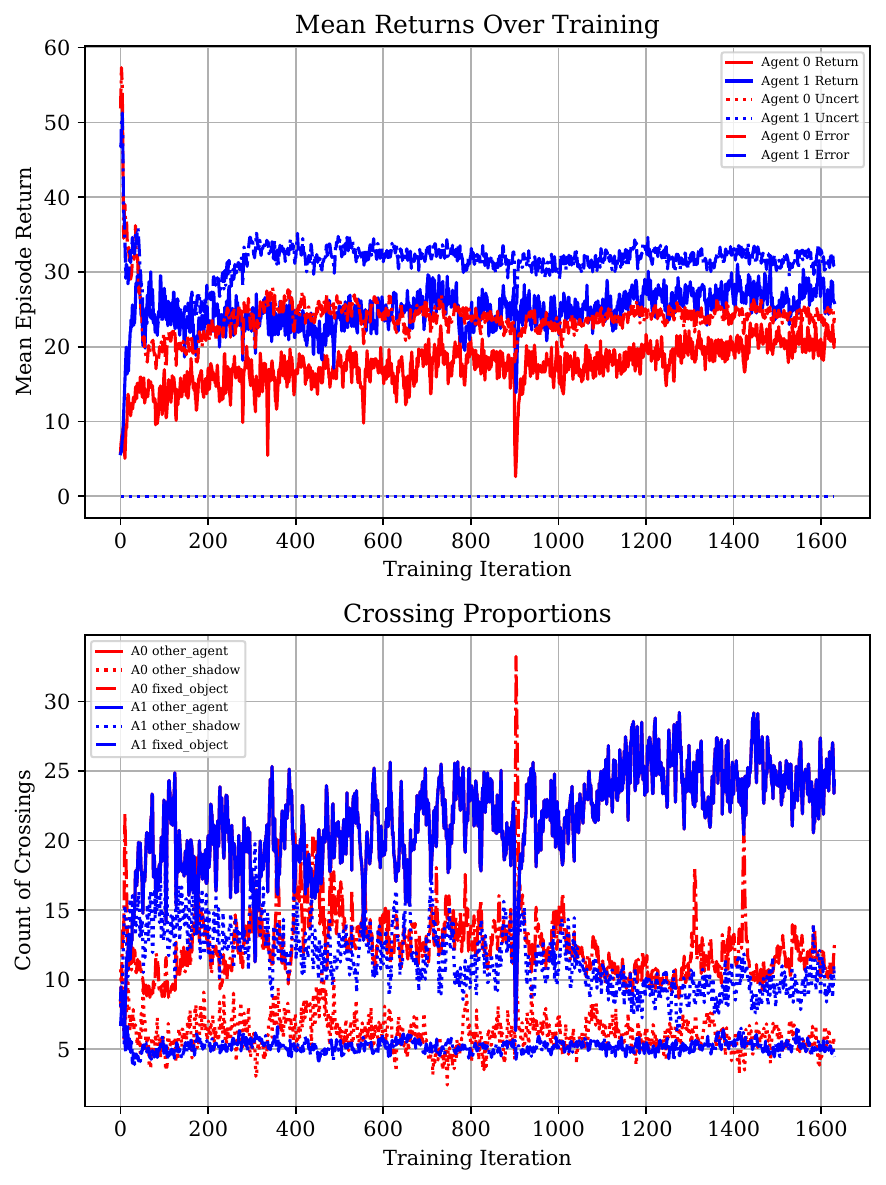}
\caption{\footnotesize Sub-reaction time anticipation: Self-other crossings are strongly preferred in this case.} 
\label{paper_plots_delay2ts}
\end{figure}

\begin{figure*}
\includegraphics[width=17cm]{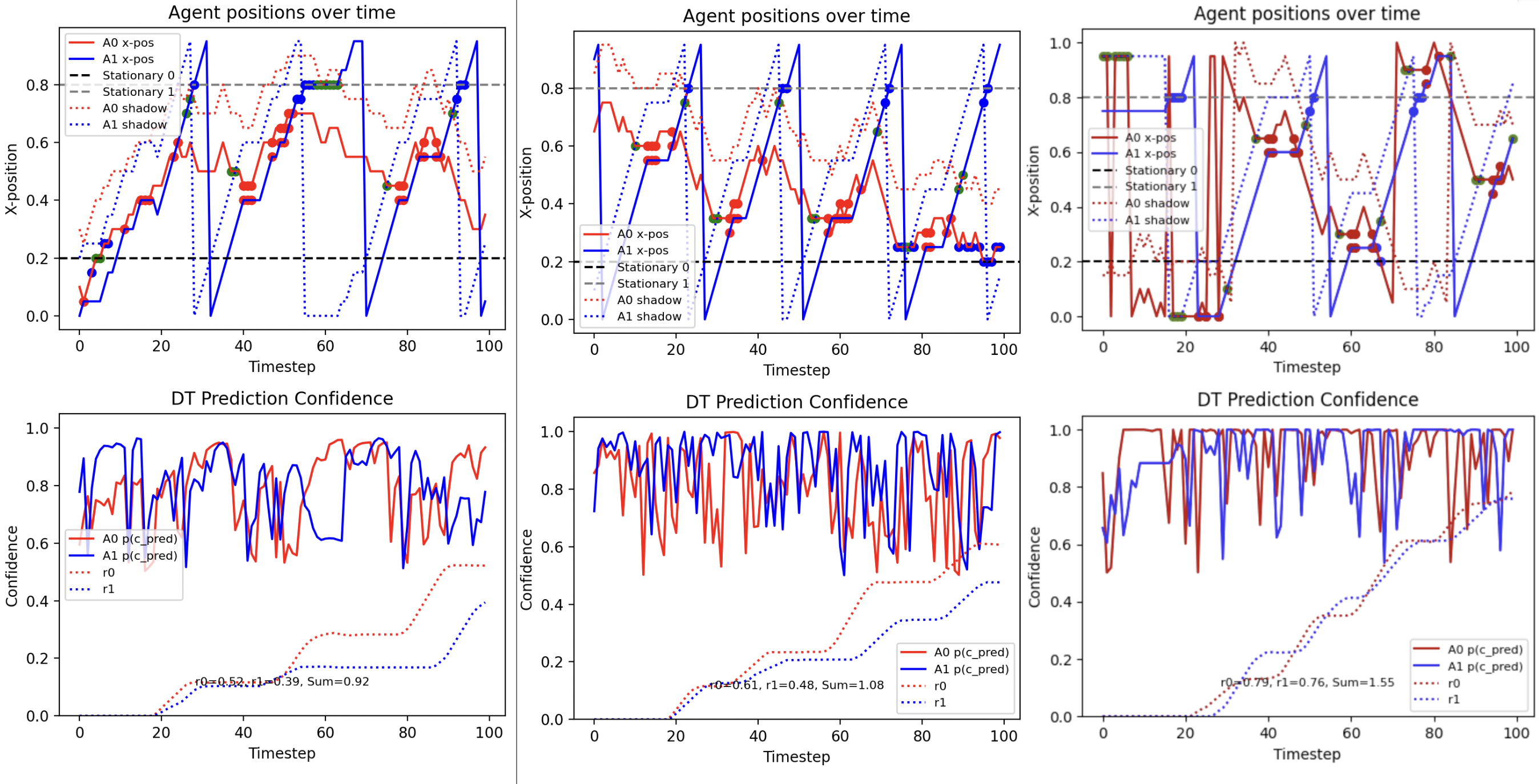}
\caption{\footnotesize Influence and Impressionability: Over the course of the run, agents learn that most reward can be achieved by imitating and being imitated by with the other. } 
\label{mi_reward3_trajectories}
\end{figure*}

We consider here only the drive for influence and impressionability. Figure \ref{paper_plots_mi_reward3} shows the returns and crossing proportions over a typical developmental trajectory, notably that self-other crossings predominate. Figure \ref{mi_reward3_trajectories} shows a sample of 3 episodes progressing through the run. Self-other association consists of periods of repeated crossing and uncrossing but with much less stereotyped patterns than before. Again, limited reward can be achieved by crossing a shadow, whilst interacting with a stationary object gives no opportunity for reward. Reward does not monotonically increase with the number of self-other crossings however; it is the mutual information that is critical, and so more systematic self-other interactions are preferred to just any random self-other interaction. \\

\begin{figure*}
\includegraphics[width=17cm]{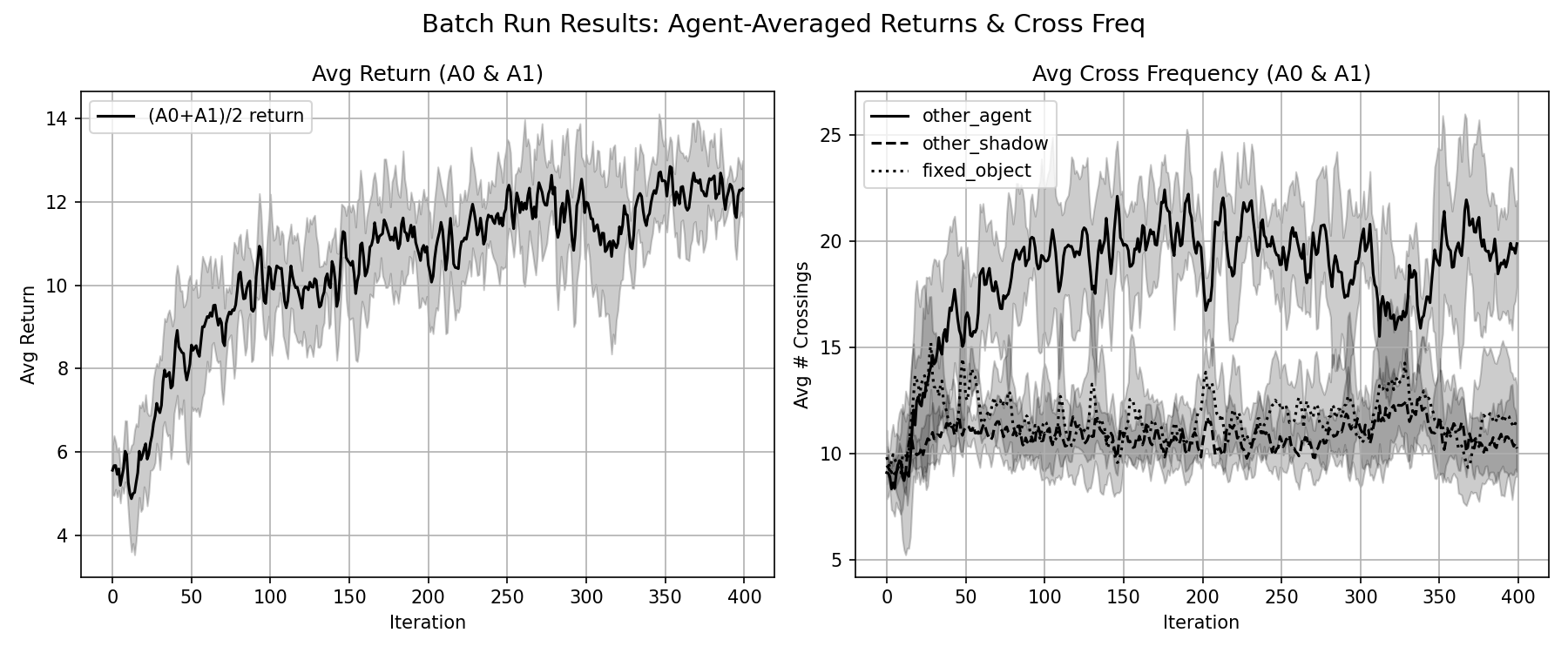}
\caption{\footnotesize In the absence of a drive for being impressionable, leaving only the drive to be imitated, self-other interaction is as prevalent as it was before.} 
\label{notListening2}
\end{figure*}

Surprisingly, Figure \ref{notListening2} shows that removing half of this drive, the drive to be impressionable, and leaving only the drive to influence does not in this case reduce the number of self-other crossings compared to where both influence and impressionability terms contribute to the reward as in Figure \ref{stats}. This is because agents learn to take it in turns to obtain reward, first agent0 influences (or is influenced by) agent1, and then agent1 influences (or is influenced by) agent0. Also, in the PCP, agent0 influencing 1 makes it more likely that agent1 can influence agent0, because they are closer together, and so a separate drive to be influenced does not seem to be needed\footnote{However, one might imagine that in a single shot non-repeated communication, or where by \textit{x} influencing \textit{y} , \textit{y} is somehow prevented from influencing \textit{x}, then both components of the drive would seem to be needed, but we leave these speculations for further work, although see results for the extrinsic reward task in Figure \ref{noListeningExtrinsic}.} \\  

\subsubsection{Sub-Reaction Time Anticipation} 

\begin{figure*}
\includegraphics[width=17cm]{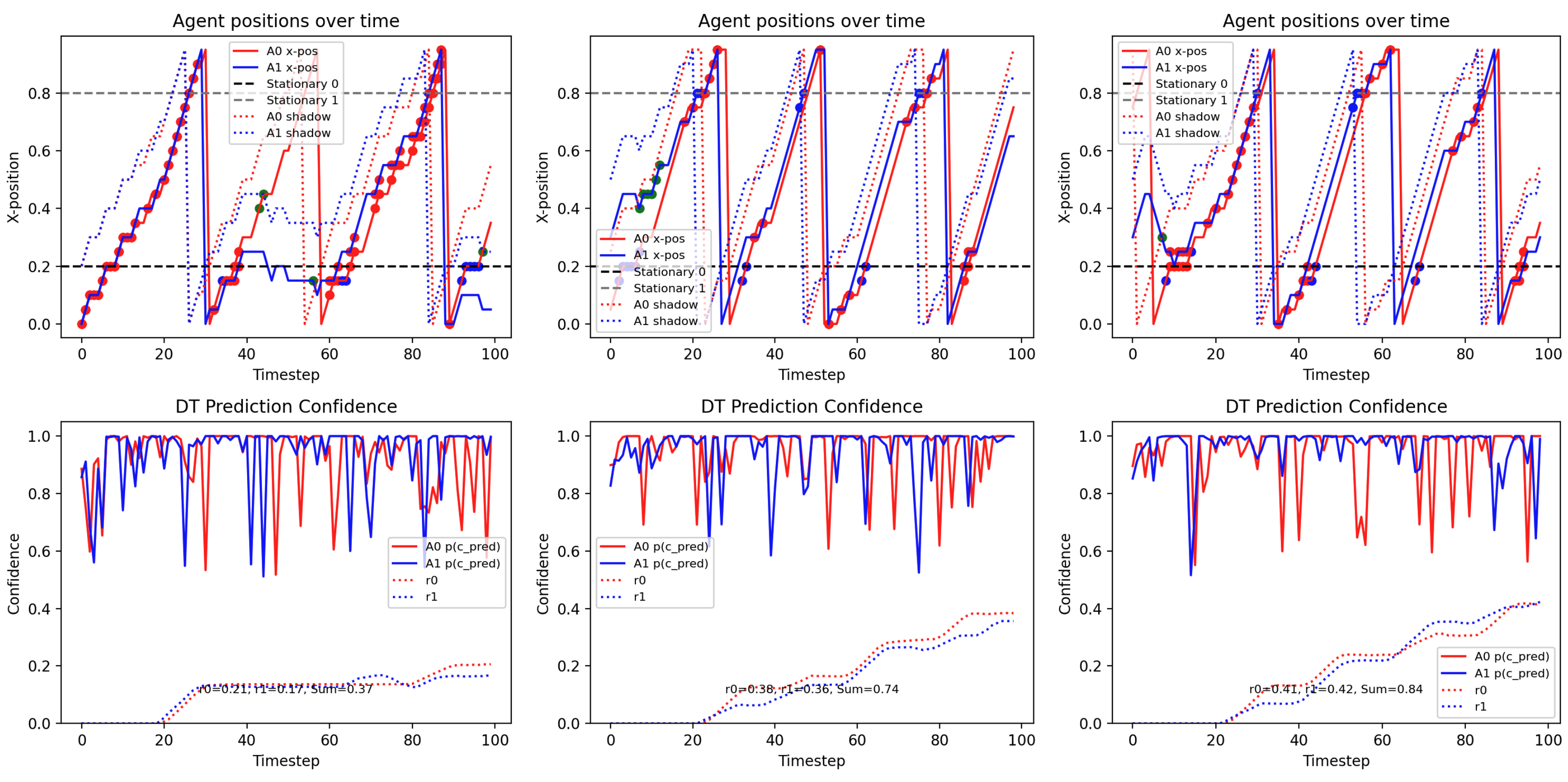}
\caption{\footnotesize Sub-reaction time anticipation: Reward is less than in with Influence and Impressionability reward without delay, however, the number of self-other crossings is greater. } 
\label{delay2ts_traj}
\end{figure*}

Even when a delay of 2 timesteps is introduced for both agents between observation and action, the same reward function can be optimized, with agents preferring self-other crossings, see Figure \ref{paper_plots_delay2ts} and Figure \ref{delay2ts_traj}. Both agents learn to move around the circle together, occasionally stopping, and catching up with the other, intermittently. 

\subsubsection{Impact on an Extrinsically Motivated Task}

\begin{figure*}
\includegraphics[width=17cm]{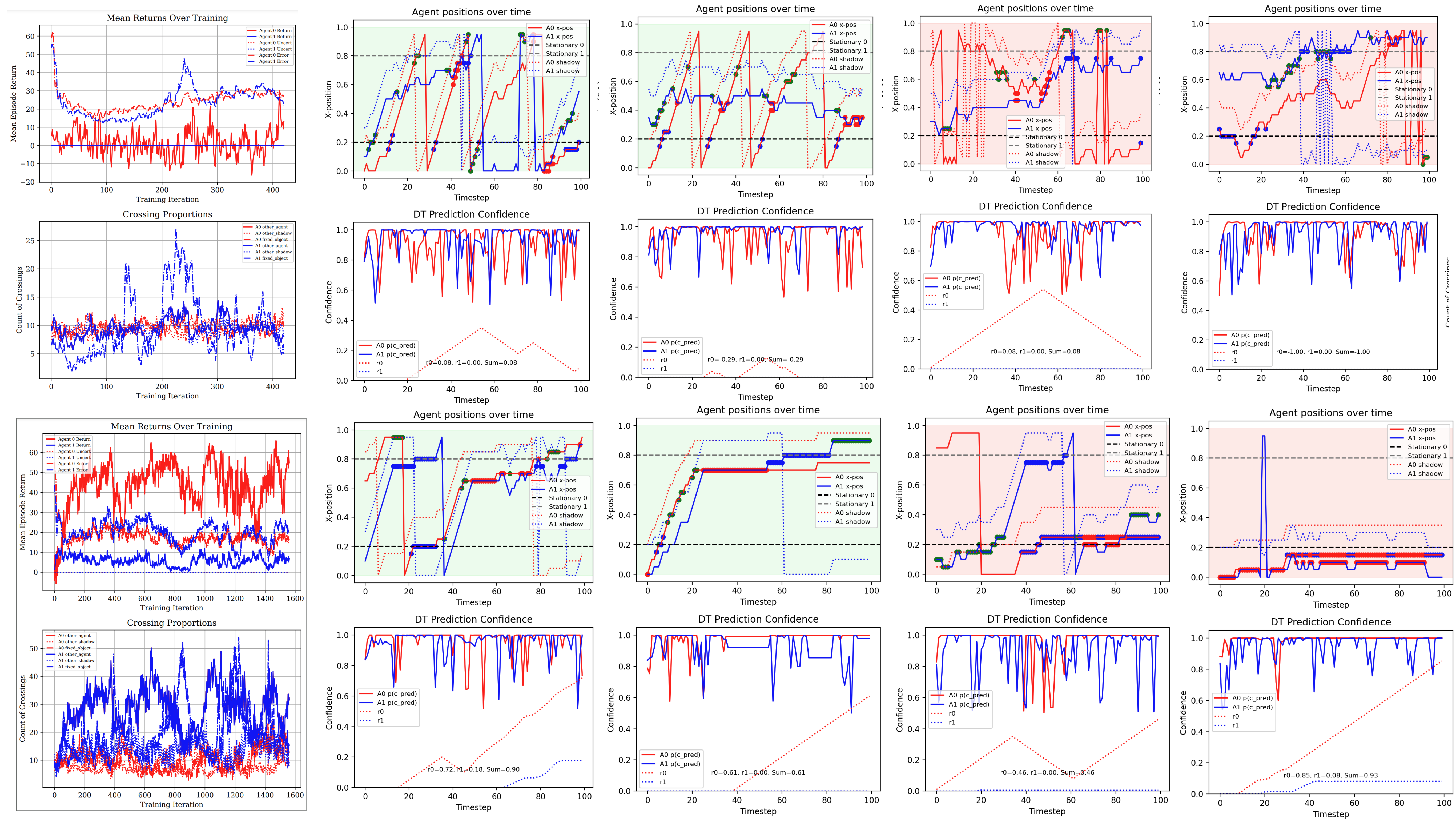}
\caption{\footnotesize The top two rows shows the training curves and 4 example episodes for the extrinsic reward task without intrinsic reward. The bottom two rows show the same, but with intrinsic reward as well. With intrinsic reward agent1 successfully goes to the top as desired by agent0 in the green (+1) tasks and to the bottom in the red (-1) tasks. No such influence is possible without the mutual information reward, and so in the top two rows agent1 does not cooperate.} 
\label{extrinsic}
\end{figure*}

Figure \ref{extrinsic} demonstrates that when the mutual information reward is coupled with an extrinsic reward given to only agent0 for signalling to agent1 that they should be in one of the two halves of the environment for a given episode, that agent0 can learn to reward agent1 through interaction, in order to manipulate/guide agent1 to the correct region. On the other hand, without intrinsic reward capability, agent1 does not cooperate. \\

\begin{figure*}
\includegraphics[width=17cm]{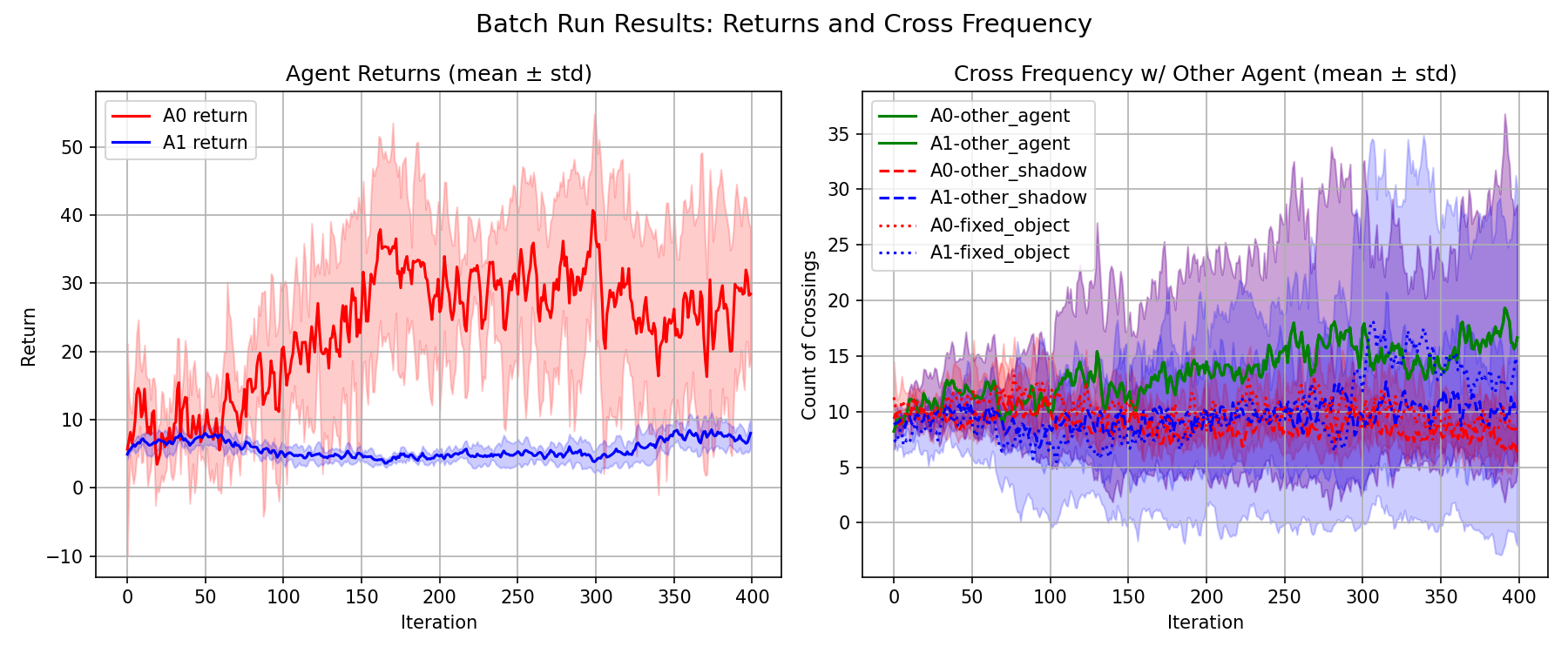}
\caption{\footnotesize Agent0's reward is in red, and Agent1's reward is in blue, averaged over 5 independent runs. Agents do not spend much time associating with each other (green line in graph on right) compared to when both components of the mutual information reward are present.}  
\label{noListeningExtrinsic}
\end{figure*}

\begin{figure*}
\includegraphics[width=17cm]{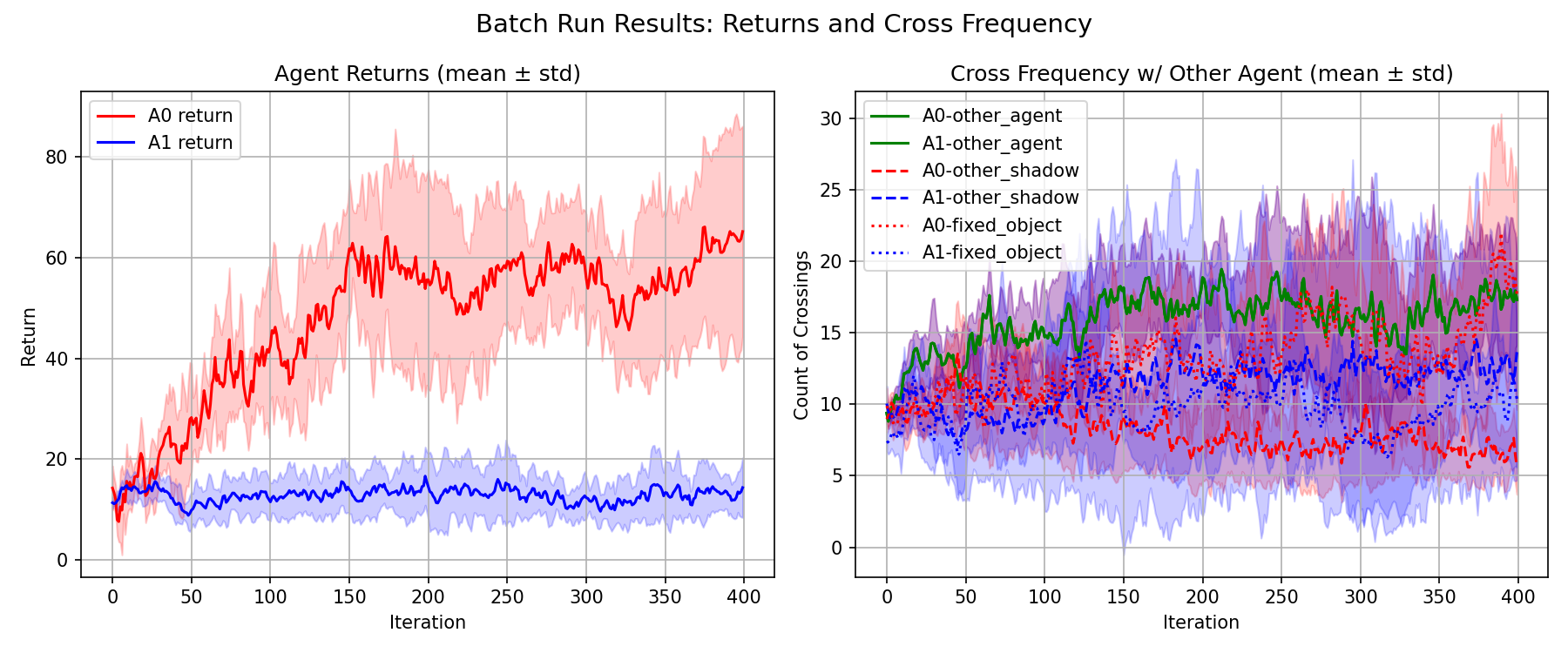}
\caption{\footnotesize Agent0's reward is in red, and Agent1's reward is in blue, averaged over 5 independent runs. In this case, the full paired influence and impressionability rewards are given.}  
\label{ListeningExtrinsic}
\end{figure*}

While impressionability rewards were not needed to establish self-other interactions (Figure \ref{notListening2}), removing them reduces performance in the extrinsic reward task, see Figure \ref{noListeningExtrinsic} compared to Figure \ref{ListeningExtrinsic} and \ref{extrinsic}. If both agents seek only influence, then agent0 finds it harder to preferentially influence agent1 to go to the correct half of the line based on agent0's private information. So in short, this investigation supports the principle that to be impressionable is useful when some extrinsic reward is to be had by being influenced by the sender (in this case the benefit is to the sender), but where influencing the sender is not necessary or desirable. Wanting to be impressionable is useful when gaining extrinsic reward depends on an asymmetric interaction where there is a learner and a teacher. 

\section{Discussion and Conclusion} 

All three examples of intrinsic motivations for coordinative engagement successfully resulted in agents spending more time interacting with each other than with inanimate objects or with each other's shadows. Artificial curiosity alone did not result in agents preferring to interact with each other. This is because it is possible to find policies for interacting with the environment that can provide surprising outcomes due to the fact that one's predictor is always limited in some way, such that a policy can always uncover its inadequacies at the boundaries of its effectiveness. However, rewards for wanting to be understood depend on reciprocal interactions which can only be achieved by interacting with another agent who has the reciprocal intention, e.g. imitating and wanting to be imitated, or influencing and wanting to be influenced, or controlling and wanting to be controlled, and this cannot be achieved typically with inanimate objects. \\

The two agents learn an other-specific policy, which is not expected to generalize to other agents. To achieve an agent-general policy we could choose 2-of-N agents to play with each other randomly at the start of each episode. This is left for further work. if this is done, it would be interesting to test the learned general policies in interaction with real humans on the PCP, if the algorithm can be implemented without significant lag. \\

The mutual information calculation used to determine influence and impressionability is somewhat contrived in this setup, meaning that for example, causal influence must be temporally tightly sequential, whereas we are happy if someone imitates us with a considerable delay. Also the features that count towards the mutual information calculation are hard coded. An alternative would be to calculate the mutual information present in the hidden layers of an inverse model of the type used in self-supervised prediction \citep{pathak2017curiosity}. The current setup assumes agents have already worked out how to tell which observations are caused by them, and which are caused by the other. MI calculation does not take into account observations which could only be caused by both agents acting together, e.g. crossing state 1 persisting although one is moving right for >2 timesteps, can only be possible if the other agent is moving with you. Whilst identifying a behavioural trajectory of this kind would be sufficient to infer a self-other type of interaction, the existing reward functions cannot detect the significance of such behaviour.\\

All the rewards considered result in the emergent property that behaviours of agent x can reward those of agent y and vice versa. This provides a rich set of ways in which agents can influence each other. When this is achieved, extrinsic tasks in which the other is not rewarded can come within the cooperative domain, because one can give the other a reward for acting in a way which would be beneficial to only oneself. In other words a new language game can be created. We have not explored the effects of mismatch between how mutual information is calculated between the two agents, but leave this for further work. \\

So far we have not allowed the intrinsic reward function to have direct access to hidden neural states. However, a relaxation of this constraint, which is entirely biologically plausible, allows for example, attractors in embeddings created in the hidden layers of a neural network by unsupervised learning processes to be available for behavioural expression, e.g. as utterances which may help the other to gain access to the insights of the discoverer. This much more direct way of communicating understanding for the benefit of the learning systems of the other will be studied in a subsequent paper, and is expected to involve mutual information calculations between performances or markings produced in the world, their referents, and the other's response to both. In the meantime we have shown a plausible outline of an intrinsic motivation function for primary intersubjectivity. \\ 



\section*{Acknowledgements}
Thanks to Tom Froese, Ben Moran, Tim Rocktäschel, Wil Cunningham, Igor Mordatch, Emily Nicholls, and Vera Vasas.

\bibliography{main}

\end{document}